\RequirePackage{fix-cm}
\RequirePackage{rotating}
\documentclass[smallcondensed, natbib]{svjour3}     
\smartqed  
\usepackage[utf8]{inputenc}
\DeclareUnicodeCharacter{04AB}{ç}
\DeclareUnicodeCharacter{02BC}{'}
\usepackage{graphics} 
\usepackage[hyphens]{url}
\usepackage{graphicx}
\usepackage{dcolumn}
\usepackage{textcomp}
\usepackage{lipsum}
\usepackage{calrsfs}
\usepackage[font=footnotesize]{caption}
\usepackage{subcaption}
\usepackage[nointegrals]{wasysym}
\usepackage{rotating}
\DeclareMathAlphabet{\pazocal}{OMS}{zplm}{m}{n}
\usepackage{epsfig} 
\usepackage{times} 
\usepackage{amsmath,amssymb} 
\usepackage{cases}
\usepackage{color}
\usepackage[fleqn]{mathtools}
\usepackage{booktabs}
\usepackage{tabulary}
\newcommand{\stt}[1]{{\small\texttt{#1}}}
\journalname{Artificial Intelligence Review}
\begin{document}

\title{Logic programming for deliberative robotic task planning
}


\author{Daniele Meli         \and
        Hirenkumar Nakawala$^*$\thanks{$^*$Research conducted while working at the University of Verona, Italy. Hirenkumar is currently affiliated with CMR Surgical Ltd, UK.} \and
        Paolo Fiorini 
}


\institute{D. Meli, H. Nakawala, P. Fiorini \at
              Department of Computer Science, University of Verona \\
              Strada Le Grazie 15, Verona, Italy \\
              \email{\{daniele.meli; hirenkumarchandrakant.nakawala; paolo.fiorini\}@univr.it}   
}

\date{\textbf{The version of record of this article, first published in Artificial Intelligence Review, is available online at Publisher’s website: \url{http://dx.doi.org/10.1007/s10462-022-10389-w}}}

\maketitle

\begin{abstract}
Over the last decade, the use of robots in production and daily life has increased. With increasingly complex tasks and interaction in different environments including humans, robots are required a higher level of autonomy for efficient deliberation.
Task planning is a key element of deliberation. It combines elementary operations into a structured plan to satisfy a prescribed goal, given specifications on the robot and the environment. In this manuscript, we present a survey on recent advances in the application of logic programming to the problem of task planning. Logic programming offers several advantages compared to other approaches, including greater expressivity and interpretability which may aid in the development of safe and reliable robots. We analyze different planners and their suitability for specific robotic applications, based on expressivity in domain representation, computational efficiency and software implementation. In this way, we support the robotic designer in choosing the best tool for his application. 
\end{abstract}


\section{Introduction} \label{sec:intro}

Robots are increasingly being used in industries, including smart industries for improving the efficiency of production environments and co-operation with the human workers \citep{bogue2011robots, bogue2018prospects}, services for the elderly and domestic assistance \citep{mandel2005towards,harmo2005needs}, and recently for autonomous driving \citep{levinson2011towards,campbell2010autonomous}. One of the most challenging requirements for robots in such complex scenarios is \emph{autonomy}, that is the ability to fulfill a \emph{mission} without external supervision. In various operating scenarios, an autonomous robot is challenged by the external state of the system, i.e., the state which is not controlled by the robot, for example unpredictability of the environment. The environment where the robot is operating is often uncertain and continuously changing, except in controlled settings as industry. Moreover, the mission may consist of one or more sub-missions named \emph{tasks}, which can be defined at run-time according to the changing situation. For instance, consider a robot operating in a domestic environment. The mission could be to search for objects located in different rooms, and move them all to a target room. This requires the completion of tasks as \emph{moving to each room, searching for one object, bringing the object to the target room}, etc. However, if an object is heavy, the mission may require the additional task of \emph{asking for human help} to carry it to destination. Furthermore, available \emph{prior knowledge} about the environment to be explored may be exploited by the robot, e.g., usual locations of objects or connections between rooms.

As a consequence, autonomy requires \emph{deliberation}, which is defined as the \emph{ability to make decisions which are motivated by reasoning on the available resources, i.e., the capabilities of the robot, the actual description of the environment and the given mission}
\citep{ingrand2017deliberation}. 

\subsection{Requirements for deliberative robotic systems}
\begin{figure}[t]
    \centering
    \includegraphics[scale=0.35]{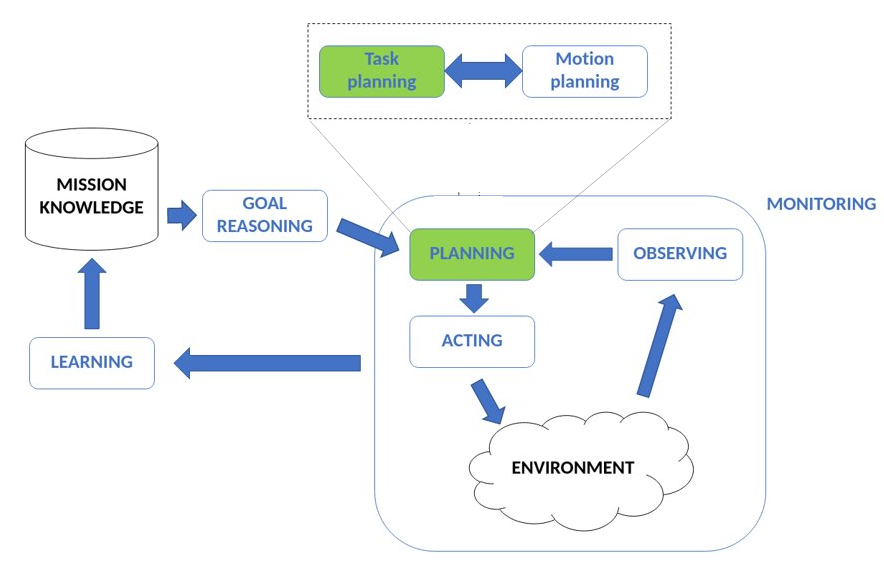}
    \caption{The functions of a deliberative robot. In this manuscript, we focus on logic-based implementations of the task planning function of a deliberative robot.}
    \label{fig:deliberative}
\end{figure}

In the last 20 years, the requirements of a deliberative robotic system have been investigated \citep{GNT16, ingrand2017deliberation, rajan2013towards, mataric2006situated}. 
Though different interpretations and characterizations of deliberation have been proposed in the field of Artificial Intelligence (AI) \citep{Bratman1987-BRAIPA, bratman1988plans, bolisani2017knowledge}, in this paper we refer to the functional model proposed by \cite{ingrand2017deliberation}, which identifies the main modules for deliberation, namely planning, acting, monitoring, observing, goal reasoning and learning (Figure \ref{fig:deliberative}).
\emph{Planning} is defined as the ability to devise a strategy to fulfill a mission, as the combination of strategies to fulfill individual tasks. 
From now on, we will refer to the strategy for a single task as the \emph{plan}, while keeping the generic nomenclature \emph{strategy} for the full mission.
The planning module considers the resources of the robot, tasks and information from the sensors, provided by the \emph{observing} module, to generate a plan. Since the planning module combines robot and environmental information, and aids in the determination of the robot's behaviour in the environment, it is a key requirement for deliberative robots. Depending on the complexity of the individual tasks, the plan can be either a motion trajectory to be directly executed by the \emph{acting} module (\emph{motion planning}), or a higher-level sequence of elementary operations to be translated into motion trajectories (\emph{task planning}). 
Reasoning on the order of execution of tasks in the mission is performed through \emph{goal reasoning}. 
Moreover, in complex scenarios where interaction with the environment and/or with humans or other robots is involved, a \emph{monitoring} module is needed to supervise the planning-acting-observing loop and guarantee the overall correctness of the execution, triggering re-planning if necessary. 
Finally, the \emph{learning} capability is required when the available information is not enough to devise a strategy for the mission. This is particularly useful, e.g., in robotics for exploration, where the environment is unknown or only partially known and information about it must be retrieved during execution. 
In this manuscript, we focus on the \emph{task planning} aspect of the deliberation process for robots.

\subsection{Task planning for deliberation}
\label{subsec:task_definition}
We now define in more detail the task planning problem considered in this paper.
Inspired by the definitions by \cite{GNT16}, we consider a \emph{robotic system} as \emph{a set of robots with a common task, which can be achieved through a set of actions, i.e., elementary operations which can be performed by the robots in the environment.}
Actions depend on task \emph{resources.}
Resources are relevant entities of the scenario, specifically robotic agents and elements of the environment (e.g., instruments which can be used in the task, or obstacles). 
The environment is defined through relevant \emph{state variables} related to specific features (e.g., position of objects and spatio-temporal relations between them).
The task planning problem is then defined by the goal, the resources and user-prescribed specifications. Specifications include \emph{preconditions} of actions, i.e., the set of conditions which must be verified before executing the actions; \emph{postconditions or effects} of actions, i.e., conditions which hold after actions are executed; \emph{execution constraints}, i.e., sets of conditions which cannot hold at the same time. Conditions involved in the specifications may be conditions on actions (e.g., an action must occur after another one) or on environmental state variables (e.g., an action can be performed only if the environment is in a specific state).
The output of the task planner is a plan, i.e., an \emph{ordered sequence of actions to be executed to accomplish the goal of the task, satisfying the task specifications}.

\subsection{Scope and structure of the paper}
While motion planning techniques have always been of interest in the robotic community from its very beginning, hence have been exhaustively surveyed for a variety of applications \citep{raja2012optimal,kamil2015review,gonzalez2015review}, a detailed and practical review on task planning for robotics is still missing.
This is mainly because task planning is an even wider problem, concerning with the coordination of actions in the long-term behavior of agents as they interact with the environment. 
This has always been of interest in the AI community \citep{ghallab2004automated, GNT16}, while the robotic community started focusing on it relatively recently, when robotic applications in more complex human scenarios have surged. 
The task planning problem can be represented and solved with a multitude of formalisms and frameworks depending on the specific application.
An overview of possible approaches is provided by several authors, e.g., by \cite{ghallab2004automated, GNT16} mainly in the context of planning for AI and with only some insights in generic task-motion planning for robotic applications; \cite{jimenez2012review} for learning-based planning; and \cite{karpas2020automated, nakawala2018approaches}, who considered generic action representation formalisms for robotics.
The purpose of our paper is to provide a comprehensive review on the state-of-the-art solutions for representing and solving the task planning problem for deliberative robots, with a focus on logic programming tools. We want to go beyond the presentation of high-level action models and formalisms, analyzing instead how these can be implemented, evaluating advantages and drawbacks of key software implementations and categorizing different planners depending on their specific field of application.
The popularity of logic programming \citep{lloyd2012foundations} has significantly increased in recent years in the robotic community, since robots are being more and more involved in human environments (as testified by recent surveys on human-robot collaboration, e.g., by \cite{goodrich2008human, chandrasekaran2015human}, and particularly safe human-robot interaction, e.g., by \cite{zacharaki2020safety, lasota2017survey}). In fact, differently from, e.g., learning-based methods \citep{goebel2018explainable}, a logic formalism allows to define the task planning problem in such a way that the autonomous behavior is \emph{interpretable} by a human supervisor, i.e., the input (typically, the environmental and intrinsic robot context, e.g., its configuration) and output (plan) to the robotic system can be easily supervised by an operator. This guarantees the reliability of the robot when it operates with other humans, e.g., in an industrial environment with other workers \citep{zhang2017plan}, or when the robot is assisting elderly or disabled people in a domestic environment \citep{alami2005task}. Moreover, interpretable planning is a requirement of the latest proposal for European regulation on high-risk AI systems\footnote{\url{https://eur-lex.europa.eu/legal-content/EN/TXT/?uri=CELEX:52021PC0206}}, and attention to explainable planning is rapidly emerging in AI and robotic research (see the work by \cite{fox2017explainable} for a deeper insight on the topic).
Focusing only on logic programming solutions for task planning allows to provide enough details for a roboticist willing to use this tool for his/her task planning problem, which would be otherwise infeasible considering the full span of related methodologies as in, e.g., \cite{GNT16}. 

In Section \ref{sec:taxonomy}, based on recent reviews, we present possible taxonomies of task planners. We then introduce a specific class of task planners based on logic programming for robotics, which offer high expressiveness in the description of the planning problem. 
Section \ref{sec:method} then specifies how the review process has been conducted, motivating the following organization of this document.
Section \ref{sec:pddl} shows how to describe the task planning problem using an action language, which constitutes the formal foundation to implement planners based on logic programming. An overview of main results of this review is then presented in Section \ref{sec:discussion}, with details provided in Sections \ref{sec:standard}-\ref{sec:probabilistic}. Finally, Section \ref{sec:conclusion} concludes the paper.

\section{Taxonomy of task planners} \label{sec:taxonomy}
The task planning problem can be described with a large number of formalisms, extensively presented in the field of AI, e.g., by \cite{GNT16,geffner2013concise}. 
Different taxonomies of task planners have been proposed.
\cite{GNT16} distinguish domain-dependent and domain-independent task planning. 
Domain-independent planning \citep{wilkins1984domain} is interesting to the robotic community because it is expected to solve the task planning problem in different environments or operational scenarios, starting from a very high-level generic description of the domain. 
For instance, in a complex robotic task involving transport of objects, the domain-independent planner is useful to define a general plan to reach the objective for all objects. 
However, manipulation of single objects depends on their specific properties and is thus more efficiently managed by a domain-dependent planner \citep{dornhege2009semantic}. 
For this reason, \cite{GNT16} propose \emph{configurable task planning} to combine high-level domain-independent task planning with domain-dependent sub-task planning. 
An explanatory example of this approach can be found in \cite{shivashankar2013godel}, where the full mission is split into a sequence of simpler tasks. 
While the global strategy is planned through domain-independent hierarchical task planning, a specific domain-dependent planner is in charge of the single tasks. 
The method is validated successfully on benchmark planning problems, including the famous block-world example. 
In general, domain-dependent planning has the potential to find more robotic applications, since it bridges the gap between formalization of the problem and practical implementation.

However, this categorization of task planners is too coarse, since it is mainly independent of the specific characteristics of the use scenario. 
An alternative classification is proposed by \cite{ingrand2017deliberation}, distinguishing between \emph{deterministic and non-deterministic task planners}. 
Non-deterministic (or probabilistic) task planners are usually based on Markov decision processes and Bayesian models. 
They capture the stochastic nature of the environment, uncertainty with the sensors and output the most probable plan \citep{kolobov2012planning,teichteil2010incremental,likhachev2005planning}.  
Probabilistic planners offer easier integration with the acting and observing modules of deliberation. 
However, they are data-driven, hence they do not generally provide enough support to the definition of complex task constraints, e.g., temporal and semantic relations between objects and robots. 

On the contrary, deterministic planners rely on a more formal description of the scenario, in terms of resources and specifications. 
Any implementation starts from a high-level action language \citep{erdem2012applications}, which represents entities of the domain and relations between them with different expressive power.
A state-of-the-art example in the field of robotics are the Planning Domain Description Language (PDDL) by \cite{aeronautiques1998pddl} and its extensions \citep{fox2003pddl2, younes2004ppddl1}.

The planning representation can then be translated to an implementative form, as proposed, e.g., by \cite{armando1999sat, vidal1996dealing, vidal1997contingent, frank2003constraint,fratini2011apsi} for temporal constraint networks with time intervals. 
Even higher expressiveness is guaranteed by planners based on logic formalisms, representing expert knowledge about the declarative task in terms of logic statements.
One key advantage of logic formalisms is the \emph{interpretability} of the generated plan when evaluated by a (possibly non technical) human supervisor. 
This guarantees the reliability of the robot when it operates with humans.
Some logic-based planners support the \emph{open-world assumption} (the truth value of a statement may be stated irrespective of whether or not it is \emph{known to be true}). 
Under the open-world assumption, description of the robotic domain can be stored in an ontology \citep{poli2010theory}, i.e., a high-level knowledge base. 
A description logic (see \cite{baader2003description} for an overview of related formalisms) then provides operators (e.g., first-order-logic operators, \cite{smullyan1995first}), allowing to reason on abstract concepts and relations encoded in the knowledge base and to instantiate them in the real world through logical inference. 
This is particularly useful for complex scenarios with a high numbers of resources, as autonomous surgery \citep{nagy2018ontology,gibaud2018toward,ICAR19}, human-robot interaction \citep{lemaignan2010oro,bruno2017caresses} and complex industrial \citep{diab2019pmk} and domestic \citep{tenorth2013knowrob} scenarios.
Description logic offers high expressivity, allowing to reason on sub-classes, properties and more complex relations between entities of the knowledge base. However, as evidenced by \cite{VG19,nakawala2019deep}, the improved expressivity also increases the computational burden, which is a major drawback in robotic applications with real-time demands. 
A more computationally tractable approach is \emph{logic programming} \citep{lloyd2012foundations}, which provides slightly reduced expressivity with respect to description logic, e.g., no quantifiers $\forall, \exists$ nor hierarchical class relations and properties. However, as we will show in the following of this paper, this still allows the application in realistic robotic domains. Moreover, in Section \ref{sec:discussion} (and sequent) we will highlight some logic programming frameworks that can be integrated with ontologies for open-world reasoning.

The categorization in deterministic and probabilistic task planners is still inadequate for a formal characterization. One drawback is that a clear decision between deterministic and probabilistic representation cannot sometimes be taken. In fact, there exist extensions of natively deterministic frameworks which support probabilistic formalism \citep{younes2004ppddl1, vidal1999handling}. Moreover, as reported by, e.g., \cite{ghallab2004automated, karpas2020automated}, planning scenarios can be better grouped into three main categories, depending on the expressivity of specifications. In this sense, it is possible to distinguish between \emph{classical planners}, assuming the robot interacts via instantaneous actions with a deterministic environment; \emph{temporal planners} which operate on complex temporal specifications; and \emph{probabilistic planners} which relax the deterministic assumption.
In this paper, we make use of this categorization for logic programming frameworks for robotic task planning. Though the distinction is not always clear (see Table \ref{tab:expressivity}), we think that the expressivity of the logic formalism is a good choice for taxonomy, since it encompasses, e.g., the distinction between deterministic and probabilistic planning.

\section{Review methodology}\label{sec:method}
\begin{table}[t]
\centering
\caption{Expressivity of the most relevant reviewed frameworks for robotic task planning with logic programming.}
\label{tab:expressivity}
\begin{tabulary}{\linewidth}{LLLL}
\toprule
& \textbf{Standard logic} & \textbf{Temporal logic} & \textbf{Probabilistic logic} \\
\midrule
SWI-Prolog \citep{WSTL12} & \checkmark & & \\
Readylog \citep{KMFS20} & \checkmark & & \\
Clingo \citep{gebser2019multi} & \checkmark & & \\
ROSPlan \citep{cashmore2015rosplan} & \checkmark & & \\ 
LTLMoP \citep{finucane2010ltlmop} & & \checkmark & \\
TuLiP \citep{WTOXM11} & & \checkmark & \\
Problog \citep{de2007problog} & & & \checkmark \\
Distributional Clauses \citep{gutmann2011magic} & & & \checkmark \\
ProbCog \citep{jain2009equipping} & & & \checkmark \\
Probabilistic CTL \citep{rutten2004mathematical} & & \checkmark & \checkmark \\
\bottomrule
\end{tabulary}
\end{table}
\begin{figure}[t]
    \centering
    \includegraphics[scale=0.2]{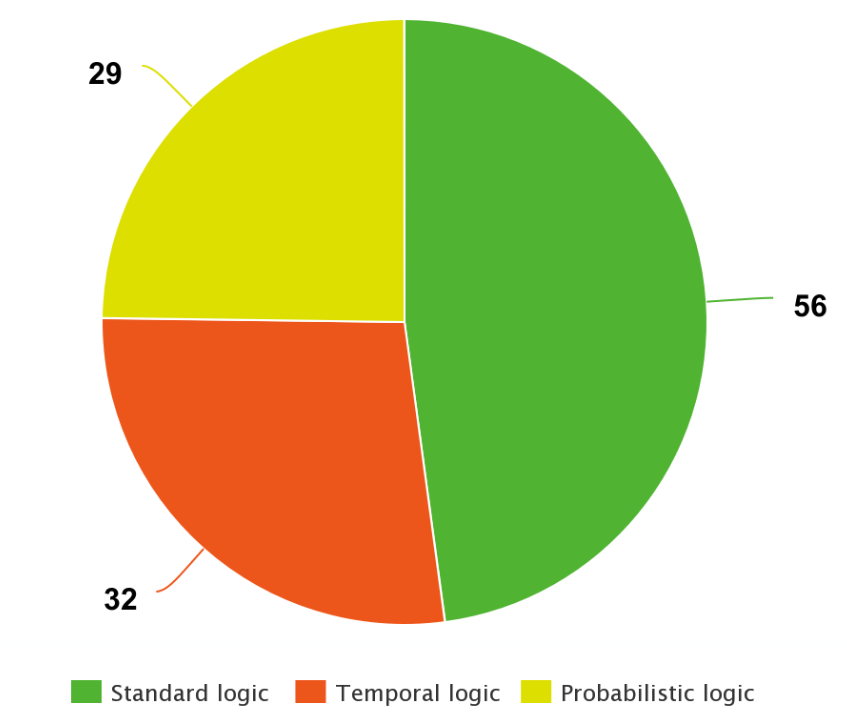}
    \caption{Categorization of logic programming frameworks and relative robotic application papers (\textbf{total 114}). 3 papers are related to probabilistic CTL, hence they are included both in the temporal and probabilistic logic category.}
    \label{fig:taxonomy}
\end{figure} 

Logic programming frameworks and relative applications appearing in this review have been selected according to a systematic search methodology.
In particular, we have searched Google Scholar and Scopus engines for keywords extracted from relevant books and reviews \citep{ghallab2004automated, GNT16, ingrand2017deliberation, mataric2006situated, BCMPN03, karpas2020automated}, such as \emph{logic programming}, \emph{task planning} and \emph{robot} (evaluated both separately and in conjunction).
Then, we have selected only frameworks and related papers with applications on real robotic systems, thus excluding, for instance, popular action representation formalisms as STRIPS \citep{fikes1971strips} and simplified action structure \citep{backstrom1991planning} - though their use in robotics is not discouraged, e.g., by \cite{karpas2020automated}.
Finally, we have decided to include only frameworks which have found concrete robotic applications in the last 20 years, in order to provide the reader with a state-of-the-art guide.
In this way, frameworks with long history (e.g., Prolog by \cite{VK76} and temporal logic programming) are still present, but their oldest robotic applications \citep{kadonoff1987prolog, tzafestas1989prolog} are discarded. 

As explained in Section \ref{sec:taxonomy}, we have organized the paper distinguishing between classical, temporal and probabilistic planning. In the context of logic programming frameworks, this results in a categorization of the underlying logic formalism, either \emph{standard Boolean logic} (classical logic operators), \emph{temporal logic} or \emph{probabilistic logic}. According to this taxonomy, Table \ref{tab:expressivity} shows the classification of relevant logic programming frameworks (i.e., $\geq2$ robotic use cases), while Figure \ref{fig:taxonomy} represents the distribution of different frameworks and relative applications in the robotic domain.

In the next section, we present the basic notions for logic programming implementation, starting from an action language description of the robotic task domain.

\section{Representing the task planning problem with logic programming}\label{sec:pddl}
Starting from a task description in human language, it is useful to represent it in an action language, in order to formally define task entities and specifications.
An action language provides the foundation for the implementation of the task planning problem in any specific logic program. 
This section shows the example formalization of the representative peg transfer task from the Fundamentals of Laparoscopic Surgery \citep{soper2008fundamentals} in the state-of-the-art action language of Planning Domain Description Language (PDDL) by \cite{PDDL}, and its translation to the syntax of a logic program.

\subsection{The peg transfer task}
\begin{figure}[t]
\centering
\includegraphics[height=5.7cm]{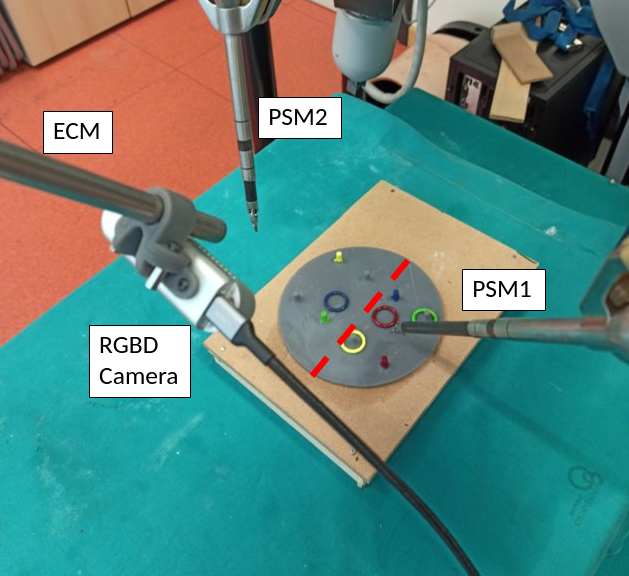}
\caption{The setup for the peg transfer task with da Vinci surgical robot. The red dashed line defines the \emph{reachability regions} for the PSMs.}
\label{fig:setup}
\end{figure}
The peg transfer (Figure \ref{fig:setup}) is a surgical training task for novice surgeons, executed on a surgical robotic systems, e.g., while learning manipulation with the state-of-the-art da Vinci surgical robotic platform (Intuitive Surgical Inc., Sunnyvale, CA, USA) \citep{dimaio2011vinci}. The setup consists of two Patient-Side Manipulators (PSMs) and an Endoscopic Camera Manipulator (ECM) with a RGBD sensor mounted on it. The PSMs, equipped with surgical tools, operate on a peg base with (up to four) colored rings, whose goal is to place them on the same-colored pegs. Several constraints influence the workflow of execution. In particular, \emph{reachability regions} for the two robotic arms are defined, in order to prevent collisions. Depending on the reachability of rings and pegs, one single arm can pick and place a ring (e.g., PSM1 with the red ring in Figure \ref{fig:setup}), or transfer from one PSM to the other may be needed (e.g., the blue ring in Figure \ref{fig:setup}). Rings may be initially placed on pegs, thus extraction may be needed before moving them to the peg or transfer point. Furthermore, colored pegs may be occupied by rings with unmatched color, hence grey pegs may be used as temporary placement for rings before task completion.

The peg transfer is chosen as a paradigmatic robotic scenario in this paper for three reasons. First, it is an instance of a pick-and-place task, so it represents a well-known use case both in the robotic and AI community. Second, the specifications of the task introduce several variations on the workflow of execution. As a consequence, enumerating sequences of actions for all possible setup configurations would not be feasible, and reasoning directly on task knowledge is a more convenient solution for planning. Third, the peg transfer is currently the benchmark in the field of Autonomous Robotic Surgery (ARS) \citep{nagy2021towards}. ARS rises safety and interpretability issues \citep{fiazza2021design} that make this task an adequate use case for the logic programming paradigms presented in this manuscript. 

\subsection{From PDDL to logic programming formalization}
Following the definitions proposed in Section \ref{subsec:task_definition}, the peg transfer task knowledge includes:
\begin{itemize}
    \item \emph{goal definition}, i.e., the rings must be placed on the same-colored pegs;
    \item \emph{resources}, i.e., rings, pegs (with the color property) and the robotic arms with possible actions. In this paper, we consider four main actions, i.e., \emph{moving to a ring} or \emph{a peg}, \emph{transferring} and \emph{extracting}\footnote{A more granular definition of actions is possible, e.g., considering gripper actions (\emph{opening} and \emph{grasping}). We decided to include them implicitly as part of the \emph{moving} actions (grasping after reaching a ring, releasing after reaching a peg), in order to keep the task description contained.};
    \item \emph{environmental variables} describing the relative positions between the resources (e.g., to evaluate reachability or peg occupancy);
    \item \emph{specifications} as reported in the previous section.
\end{itemize}
PDDL\footnote{Here, PDDL 3.1 syntax is used as of \url{https://helios.hud.ac.uk/scommv/IPC-14/repository/kovacs-pddl-3.1-2011.pdf}.} allows to represent this knowledge in a formalism which specifies a \emph{domain definition} and a \emph{problem instance}.
The domain definition contains declaration of \emph{types}, i.e., classes of objects in the resources (e.g., objects of type \stt{Ring} or \stt{Peg}); declaration of names and argument types of \emph{predicates} representing specific features of the domain (e.g., \stt{on(Ring, Color, Peg, Color)} for relative positions of rings with respect to pegs); \emph{actions} with their parameters, preconditions and effects. The domain definition for peg transfer is as follows:
\begin{subequations}
    \begin{align*}
        \stt{(define}&\stt{(domain }\stt{peg\_transfer)} \\
        &\textit{\% defining equality (=) between entities and allowing quantifier in goal}\\
        &\stt{(:requirements :equality, :universal-precondition)}\\
        &\stt{(:types } \stt{Ring, Peg, Arm, Color)} \\
        &\stt{(:predicates }\stt{(holding ?a, ?r, ?c - Arm, Ring, Color);} \\
        &\ \ \ \ \ \ \ \ \ \ \stt{(on ?r, ?c1, ?p, ?c2 - Ring, Color, Peg, Color)} \\
        &\ \ \ \ \ \ \ \ \ \ \stt{(reachable ?a, ?p, ?c - Arm, Peg, Color)} \\
        &\ \ \ \ \ \ \ \ \ \ \stt{(reachable ?a, ?r, ?c - Arm, Ring, Color)} \\
        &\stt{(:action }\stt{move} \\
        &\ \ \ \ \ \ \ \ \ \ \stt{parameters: (?a - Arm; ?r - Ring; ?c - Color)} \\
        &\ \ \ \ \ \ \ \ \ \ \stt{precondition: (reachable ?a, ?r, ?c)} \\
        &\ \ \ \ \ \ \ \ \ \ \stt{effect: (holding ?a, ?r, ?c))}\\ 
        &\stt{(:action }\stt{move}\\
        &\ \ \ \ \ \ \ \ \ \ \stt{parameters: (?a - Arm; ?p - Peg; ?c1 - Color)} \\
        &\ \ \ \ \ \ \ \ \ \ \stt{precondition: (and (reachable ?a, ?p, ?c1)} \\
        &\ \ \ \ \ \ \ \ \ \ \ \ \ \ \ \ \ \ \ \ \ \ \stt{(holding ?a, ?r, ?c2)}\\
        &\ \ \ \ \ \ \ \ \ \ \ \ \ \ \ \ \ \ \ \ \ \ \stt{(not (on ?r, ?c3, ?p, ?c1))))}\\
        &\ \ \ \ \ \ \ \ \ \ \ \ \ \ \ \ \ \ \ \ \ \ \stt{(not (on ?r, ?c2, ?p, ?c4))))}\\
        &\ \ \ \ \ \ \ \ \ \ \stt{effect: (on ?r, ?c2 ?p, ?c1))}
    \end{align*}
\end{subequations}
\begin{subequations}
    \begin{align*}
        &\stt{(:action }\stt{extract} \\
        &\ \ \ \ \ \ \ \ \ \ \stt{parameters: (?a - Arm; ?r - Ring; ?c1 - Color)} \\
        &\ \ \ \ \ \ \ \ \ \ \stt{precondition: (holding ?a, ?r, ?c1)}\\
        &\ \ \ \ \ \ \ \ \ \ \stt{effect: (not (on ?r, ?c1, ?p, ?c2)))}\\
        &\stt{(:action }\stt{transfer} \\
        &\ \ \ \ \ \ \ \ \ \ \stt{parameters: (?a1 - Arm; ?r - Ring; ?c1 - Color)} \\
        &\ \ \ \ \ \ \ \ \ \ \stt{precondition: (and (holding ?a1, ?r, ?c1)}\\
        &\ \ \ \ \ \ \ \ \ \ \ \ \ \ \ \ \ \ \ \ \ \ \stt{(not (on ?r, ?c1, ?p, ?c2)))}\\
        &\ \ \ \ \ \ \ \ \ \ \stt{effect: (and (holding ?a2, ?r, ?c1)}\\
        &\ \ \ \ \ \ \ \ \ \ \ \ \ \ \ \ \ \ \ \ \ \ \stt{(not (holding ?a1, ?r, ?c1))}\\
        &\ \ \ \ \ \ \ \ \ \ \ \ \ \ \ \ \ \ \ \ \ \ \stt{(not (= (?a1, ?a2)))))}
     \end{align*}
\end{subequations}
Given the domain definition, a problem instance specifies the \emph{goal} of the task; the initial state \emph{init}, i.e., initial values for predicates; \emph{objects}, i.e., instances of types (e.g., actual rings and pegs with specific colors). One possible problem instance for peg transfer is as follows:
\begin{subequations}
    \begin{align*}
        \stt{(define}&\stt{(problem }\stt{peg\_transfer\_instance)} \\
        &\stt{(:domain }\stt{peg\_transfer)} \\
        &\stt{(:objects } \stt{(ring - Ring)} \\
        &\ \ \ \ \ \ \ \ \ \ \stt{(peg - Peg)} \\
        &\ \ \ \ \ \ \ \ \ \ \stt{(blue - Color)} \\
        &\ \ \ \ \ \ \ \ \ \ \stt{(psm1 - Arm)} \\
        &\ \ \ \ \ \ \ \ \ \ \stt{(psm2 - Arm))} \\
        &\stt{(: init (reachable psm1, ring, blue) (reachable psm2 peg blue))}\\
        &\stt{(:goal (forall (?c - Color; ?r - Ring)}\\
        &\ \ \ \ \ \ \ \ \ \ \stt{(and (on ?r, ?c ?p, ?c)}\\
        &\ \ \ \ \ \ \ \ \ \ \stt{(reachable ?a, ?r, ?c)))))}
    \end{align*}
\end{subequations}
Starting from this high-level representation, a logic programming language encodes the task knowledge using a \emph{sorted signature} $\mathcal{D}$, with symbols (\emph{terms)} which can be arranged hierarchically to form \emph{atoms}, i.e., predicates of terms representing attributes of the domain. Terms may be variables or constants. A term which is constant is \emph{ground}, and atoms are ground when terms in them are all ground. Atoms are either \emph{fluents} or \emph{statics}, depending on whether they depend on time or not, respectively\footnote{As clarified in the next sections, dependency on time can be made explicit through the use of a temporal variable, depending on the specific logic programming framework.}.

In order to clarify these concepts, consider the peg transfer task description in PDDL. \stt{Ring}, \stt{Peg}, \stt{Color} and \stt{Arm} are static variables, with possible constant values defined by objects in problem instance. From now on, we will use the short notation \stt{R, P, C, A} when referring to types as unassigned variables, respectively.
Fluents are environmental features (PDDL predicates) with their parameters, i.e., \stt{reachable(A, P/R, C)}, \stt{on(R, C1, P, C2)}, \stt{holding(A, R, C)}, and actions \stt{move(A, R/P, C)}, \stt{extract(A, R, C)}, \stt{transfer(A, R, C)}.

Logic programming languages also introduce \emph{operators} which allow to write the specifications. 
Different operators are allowed, e.g., temporal (e.g., \emph{until} and \emph{release} operators) or Boolean (e.g., logic implication and conjunction), depending on the underlying logic formalism. This will be clarified in the next sections.
For instance, consider \stt{move(A, P, C)} in the PDDL description for peg transfer. Preconditions and effects can be represented as implications between atoms:
\begin{align*}
    \stt{move(A, P, C1)} &\leftarrow \stt{reachable(A, P, C1)} \land \stt{holding(A, R, C2)} \land\\ &\ \ \ \ \ \ \lnot\stt{on(R, C3, P, C1)} \land \lnot\stt{on(R, C2, P, C4)}\\
    \stt{on(R, C2, P, C1)} &\leftarrow \stt{move(A, P, C1)}
\end{align*}
where the left-hand side is usually referred to as \emph{head}, while the right-hand side as \emph{body} of the specification.
Similarly, it is possible to define constraints in the form $\bot \leftarrow$ \stt{atoms}, being $\bot$ the logic \emph{falsum}. For instance, the precondition for \stt{move(A, P, C)} can be also expressed as:
\begin{align*}
    \stt{move(A, P, C1)} &\leftarrow \stt{reachable(A, P, C1)} \land \stt{holding(A, R, C2)}\\
    \bot &\leftarrow \stt{on(R, C3, P, C1)} \land \stt{move(A, P, C1)}\\
    \bot &\leftarrow \stt{on(R, C2, P, C4)} \land \stt{holding(A, R, C2)} \land \stt{move(A, P, C1)}
\end{align*}
Given a logic program, a language-specific \emph{solver} computes the \emph{plan} as the set of ground actions, starting from initial grounding of environmental variables from available evidence.  
Grounding translates the program to a set of \emph{propositional logic formulas}, where grounded atoms become Boolean variables.
Then, logic deduction is exploited to propagate grounded atoms through specifications, until a plan is found. 
For example, consider the above PDDL problem instance.
Initial conditions satisfy preconditions for \stt{move(psm1, ring, blue)}. This leads to the grounding of corresponding effect \stt{holding(psm1, ring, blue)}. Then, precondition for \stt{transfer(psm1, ring, blue)} holds, whose effect leads to grounding of \stt{holding(psm2, ring, blue)} and, finally, last action \stt{move(psm2, peg, blue)} to complete the problem instance.

Before detailing how representation and solving specialize for different logic programming frameworks, we give an overview of main results of this review, in order to guide the interested reader to a deeper insight only in the relevant sections for his/her specific needs.
\begin{sidewaystable}
\centering
\caption{Summary of the main reviewed frameworks for robotic task planning through logic programming. Legend for entries: $\mathcal{F}$ = formalism (\emph{Pl} = Prolog); \emph{ROS} = ROS support; \emph{Impl.} = software implementation; \emph{Integrated} specifies whether the framework is stand-alone or is integrated with other modules; \emph{Online} = online adaptation / plan revision. Legend for applications: NAV=navigation; S$\&$R=search$\&$rescure; DOM=domestic environment; IND=industry, SRV=service; HRI=human-robot interaction, MRC=multi-robot coordination; MAN=manufacturing; SUR=surveillance; LOC=locomotion, MANIP=manipulation.}
\label{tab:summary}
\begin{tabulary}{\textwidth}{llCCCCC}
\toprule
& $\mathcal{F}$ & \textbf{Applications} & \textbf{Impl.} & \textbf{ROS} & \textbf{Integrated} & \textbf{Online}\\
\midrule
SWI- Prolog \citep{WSTL12} & Pl & DOM, SRV, NAV, IND, S$\&$R & C++, Java, Python & \checkmark & In CRAM \citep{beetz2010cram} (open-world), Tartarus \citep{SBSJN15} (MRC) & \checkmark\\
\midrule
Readylog \citep{KMFS20} & Pl & SRV, DOM & C++ & \checkmark & & \checkmark\\
\midrule
Clingo \citep{gebser2019multi} & ASP & DOM, SRV, IND & C, Python & \checkmark & & \checkmark\\
\midrule
ROSPlan \citep{cashmore2015rosplan} & PDDL & NAV, HRI, MRC & & \checkmark & &\\ \midrule
LTLMoP \citep{finucane2010ltlmop} & LTL & NAV, S$\&$R, HRI, MRC & Python & \checkmark & & \\
\midrule
TuLiP \citep{WTOXM11} & LTL & SUR, LOC & Python & & Hybrid controller design & \\
\midrule
Problog \citep{de2007problog} & DS & MANIP & Python, Java & & & \checkmark\\
\midrule
DC \citep{gutmann2011magic} & DS & MANIP & & & & \checkmark\\
\midrule
ProbCog \citep{jain2009equipping} & BLN & DOM, MANIP, NAV & Python, Java & \checkmark & Statistical learning$\&$reasoning & \checkmark\\
\midrule
PCTL \citep{rutten2004mathematical} & CTL & MAN, S$\&$R, MRC & & & \\
\bottomrule
\end{tabulary}
\end{sidewaystable}

\section{Overview of logic programming frameworks for robotic task planning}\label{sec:discussion}
A summary of the results of the analysis of the state of the art in logic programming for robotic task planning is presented in Table \ref{tab:summary}. 
The choice of the appropriate logic programming framework for task planning depends on several criteria, which will be further highlighted in the end of Sections \ref{sec:standard}-\ref{sec:probabilistic}:
\subsection{Expressivity of logic formalism}
    The \emph{expressivity} of the underlying logic formalism must be considered. Our review has identified several applications of logic programming for robotic task planning, including domestic/service scenarios, navigation, industry, search $\&$ rescue, human-robot interaction, multi-robot coordination, surveillance and manipulation tasks. Most of these applications are covered by standard logical planners, which allow to specify conditional laws, constraints and optimization statements for optimal plan generation. The state-of-the-art Prolog-based planner, SWI-Prolog \citep{WSTL12} ($>250$ citations on Scopus), can cope with different data types, including real variables, supporting the designer in the definition of proper specifications for real robots dealing with continuous data from sensors. Planners based on the answer set semantics are still limited in this sense, though the main representative Clingo by \cite{gebser2019multi} ($>50$ citations on Scopus) allows to define arithmetic constraints. 
    Moreover, Clingo supports multiple plan computation and preference reasoning for optimal planning.
    
    Temporal planners shall be preferred when complex temporal relations between robotic agents and the environment are required to properly describe the scenario. This is the case, for instance, of tasks requiring long human-robot interaction, or (multi-robot) exploration of large complex environments. 
    Moreover, temporal logic can easily express conditions from system theory, as invariance and reachability of sets (e.g., for definition of system stability). This makes planners based on temporal logic advantageous for applications with a tight integration of task and motion planning, e.g., multi-robot exploration. To this aim, TuLiP toolbox \citep{wongpiromsarn2010receding} ($>100$ citations on Scopus) is the state-of-the-art solution. However, it is fair to remark that some temporal concepts can be expressed with standard logical planners, e.g., using event calculus in answer set semantics as in \cite{son2006domain, meli2021inductive}, at the expense of increased size of represented knowledge (e.g., additional temporal variables or specifications). 
    Moreover, temporal extensions of answer set programming have been recently introduced, e.g., \emph{telingo} \citep{cabalar2019telingo}, though not applied to robotic use cases yet.
    Hence, the choice of the logic formalism should be tailored on the peculiarities of the applicative domain and the most relevant and frequent concepts (e.g., either temporal or causal) to be represented.
    
    Probabilistic planners are useful in tasks involving continuous interaction with the environment, requiring reasoning on uncertain sensor information to refine the knowledge of the scenario. This may be the case, for instance, of manipulation and navigation tasks. 
    Main implementations are Problog \citep{de2007problog} ($>450$ citations on Scopus), which allows to reason on discrete probability distributions with a syntax similar to Prolog, and Distributional Clauses (DC) by \cite{gutmann2011magic} ($\approx50$ citations on Scopus), representing continuous probability distributions.
\subsection{Computational efficiency}
    An important requirement is the \emph{computational efficiency} of plan generation.
    This is related to the expressivity of the underlying logic formalism.
    In fact, in general the problem of deciding whether a problem is solvable is PSPACE complete in the case of classical planning \citep{bylander1994computational}, and becomes EXPSPACE complete when considering temporal planning \citep{rintanen2007complexity}, or even undecidable when considering continuous variables, e.g., in probabilistic planning \citep{helmert2002decidability}.
    As a consequence, it is important to carefully choose the most suitable logic programming framework for the specific robotic application, identifying the best tradeoff between expressivity and efficiency.
    In general, standard logic programming frameworks should be preferred, also because they implement most efficient algorithms for solving, and allow to represent complex task planning domains with realistic number of robotic agents, actions and environmental features.
    In particular, planners based on the answer set semantics offer some advantages with respect to Prolog-based, e.g., they do not require prior stratification of the task knowledge and the ordering of statements and atoms in them is not relevant. 
    
    On the contrary, temporal and probabilistic logic programming framework should be considered only when the complexity (number of resources and specifications) of the scenario is limited. In fact, the solving efficiency of these programs does not scale well with the problem dimension, due to state explosion problems, unless the specific use case justifies strict assumptions, e.g., on the limited time horizon for the plan or neglectible specifications.
    When advanced expressivity is required, currently the best solution is to rely on frameworks which combine standard logic programming framework with specific tools, e.g., answer set semantics in conjunction with Markov decision processes \citep{SGZW19} or \emph{telingo} \citep{cabalar2019telingo} to represent temporal operators in Clingo. 
    
\subsection{Software implementation}
    \emph{Software implementation and programming language support} are important factors to choose a logic programming framework, in order to build robotic architectures integrating multiple deliberative functions. 
    Standard logic programming is well established in the robotic community, hence it offers integration with the standard Robot Operating System (ROS) \citep{koubaa2017robot} for robotic research. ROSPlan \citep{cashmore2015rosplan} ($>150$ citations on Scopus) is a particularly versatile framework, since it does not require any specific logic formalism for task specifications, but generic PDDL interpretations. 
    Clingo and Prolog also offer APIs for integration in ROS-compliant languages.
    
    The only temporal planner logic programming framework offering direct integration with ROS is LTLMoP \citep{finucane2010ltlmop} ($>100$ citations on Scopus), while ProbCog \citep{jain2009equipping} ($\approx25$ citations on Scopus) was born as a probabilistic extension to the CRAM project \citep{beetz2010cram}, hence it offers built-in ROS interface. However, relevant frameworks as TuLiP and Problog support Python and Java, hence they can be easily integrated in the ROS framework.
    
\subsection{Support for open world planning}
    \emph{Support for open-world planning} is an interesting feature for complex robotic applications, e.g., exploration of large environments with many resources, where external knowledge bases may be queried in order to provide useful information about the domain. 
    In general, Prolog-based tools offer the best performance in these scenarios, since they natively support query-based reasoning.
    SWI-Prolog is part of the well established CRAM project for open-world planning, but also novel DLVHEX \citep{eiter2018dlvhex} for the answer set semantics offers easy integration with external knowledge bases, though applications to robotics are not known at the writing of this survey. 
    Among probabilistic planners, ProbCog is the natural statistical extension to the CRAM project.
    
\subsection{Plan revision opportunity}
    Finally, \emph{plan revision} is a key ability for online robotic task planners, since it allows to cope with dynamic conditions which occur in many robotic scenarios. 
    Planners based on Prolog and the answer set semantics rely on the formalism of non-monotonic logic, hence they inherently support online plan adaptation. 
    
    Also several probabilistic planners allow plan revision, since they are usually stochastic extensions of Prolog-based planners. Most temporal planners do not offer this feature, because they are based on monotonic versions of LTL. This is partly due to the different scope of temporal planners, which are intended to generate correct-by-constructions plans. 

\section{Standard logic programming}\label{sec:standard}The terminology \emph{standard logic} refers to logic programming frameworks based on classical Boolean logic, where specifications are expressed using operators as negation, conjunction, disjunction and implication.
In order to represent the temporal sequence of actions and fluents, an explicit time variable \stt{t} must be defined. Hence, with reference to the peg transfer domain, fluents have an additional argument \stt{t} (e.g., \stt{on(R, C1, P, C2, t)}). Given this formalism, it is important to introduce an explicit temporal delay between actions and their effects, e.g., for \stt{move(A, P, C, t)} in the peg transfer task:
\begin{equation*}
    \stt{on(R, C1, P, C2, t+1)} \leftarrow \stt{move(A, P, C2, t)}
\end{equation*}
In this way, a temporal sequence of actions can be generated.

Logic programming frameworks in this class can be grouped in two main categories, either based on Prolog (Section \ref{subsec:prolog}) or answer set semantics (Section \ref{subsec:asp_review}), which differ mainly in the solving strategy.
Other application-specific implementations are mentioned in Section \ref{subsec:others}.

\subsection{Prolog-based planners}
\label{subsec:prolog}
One of the first examples of standard logic programming languages for planning is \emph{Prolog}, introduced in \cite{VK76}. Specifically, Prolog is a language for \emph{non-monotonic logic programming}, i.e., the underlying logic is non-monotonic \citep{mcdermott1980non}. 
To understand non-monotonicity, consider a generic task knowledge $K$ expressed as a set of goal, resources and specifications, and a Boolean assertion $A$ entailed by $K$, i.e., $K \models A$. The knowledge is said to be monotonic if the addition of new knowledge $S$ (e.g., further specifications or new evidence) does not affect the truth of the entailment, i.e., $K \cup S \models A$. Otherwise, the knowledge is said to be non-monotonic. Non-monotonicity of knowledge has allowed to solve the famous qualification problem in AI, moving from the concept of circumscription of human reasoning \cite{MccaJ80}, and it has led to the development of several logic frameworks for decision making of expert systems, e.g., deductive databases \cite{ELMPS97}, abduction theory \cite{ALPQ00} and stable model semantics \cite{DNK97}. In robotic tasks, planning under non-monotonicity assumption is crucial, since the environment is usually not completely known or dynamic. Considering the peg transfer task, in order to emulate the anatomical variability in the surgical context, rings may be moved in the scene (e.g., by a human) during task execution, thus altering the reachability conditions. As a consequence, the solver must continuously update the plan and prescribe, e.g., transfer between arms instead of direct placement on peg.

The syntax of Prolog extends standard logic formalism with \textit{negation as failure} (NAF). To show the difference between NAF and classical logic negation, consider a generic Boolean variable $p$. In order to state $\lnot p$ in classic logic, it is necessary to prove that the negation of $p$ holds; on the contrary, to prove \stt{not(p)} (NAF syntax in Prolog) it is only required that $p$ does not hold. The function of NAF in Prolog is two-fold. First, it introduces the \emph{closed-world assumption}, since it is treated in the solving process as a classical negation (in other words, \emph{unknown facts are assumed to be false}). Second, NAF supports the non-monotonic paradigm, since it is possible to specify that an atom does not hold, and then revise it as new evidence is acquired. For instance, the goal for the peg transfer task can be expressed as the constraint:
\begin{equation*}
    \stt{:- reachable(A, R, C, t), not(on(R, C, P, C, t))}. 
\end{equation*}
where \stt{:-} represents $\leftarrow$ in standard Boolean logic (in this specific case, it equals to $\bot \leftarrow$, since $\bot$ is omitted in standard Prolog syntax).
This specification is not satisfied until \stt{on(R, C, P, C, t)} is not grounded for any reachable ring (i.e., until all rings in the scene are not placed on the same-colored pegs). \stt{on(R, C, P, C, t)} can be grounded either from external sensor information, or from specifications (e.g., effect of \stt{move(A, P, C, t)} action).

Starting from Prolog III in \cite{ColmA90} and up to the latest version Prolog IV in \cite{NarbG99}, Prolog has further extended the standard logic syntax with structures like lists and trees to represent a richer hierarchy between atoms, and is able to manage constraints on real variables which are useful for robotic applications involving continuous environmental and kinematic variables (e.g., 3D positions of rings and pegs for peg transfer task). 

These features have increased the appeal of Prolog in many challenging robotic scenarios. Relevant are the many Prolog-based implementations of the Golog action language \citep{levesque1997golog}, such as in \cite{JVBS16} for multi-robot hierarchical task planning in the context of RoboEarth project \citep{waibel2011roboearth}; \cite{KMFS20}, where a useful integration of the Golog dialect Readylog with ROS is proposed through C++ bindings (Golog++ by \cite{MSF18}), with an application to the Pepper service robot; \cite{GNCL16}, where INTRGOLOG is presented to implement task interruption and resumption in reaction to anomalous events; \cite{SFL12}, where Readylog is used for task planning in domestic scenario with Caesar robot; \cite{FFL07}, where TEAMGOLOG is used for multi-robot coordination in a search-and-rescue application under partial observability. Among main implementations of Prolog, also SWI-Prolog \citep{WSTL12} shall be mentioned, which implements constructs for optimal query answering and plan generation under the paradigm of preference reasoning \citep{BNT08} and has been used, e.g., in the Tartarus framework for the integration and coordination of multiple robots in cyber-physical systems \citep{SBSJN15}; \cite{MW07}, where Fuzzy Prolog is used to deal with missing information in real-time robotic soccer; \cite{JC17}, where multi-robot coordination for domestic activities is implemented in Prolog; \cite{XWZTZ14}, where an efficient software integration between qualitative Prolog calculus and quantitative C++ processing is implemented for robotic assistance to elderly and disabled people; \cite{PSMRF13}, which proposes SitLog, a Prolog-based planner for service robotic tasks in domestic scenario, in the context of RoboCup@Home international competition; \cite{NTM08, FMLDP16} for robotic assembly in industry.

As detailed in \cite{NarbG99}, Prolog represents task knowledge as a tree structure, following the variable-atom and atom-operator hierarchy. 
For instance, consider the statement representing the precondition of \stt{move(A, R, t)} in the peg transfer task knolwedge:
\begin{equation*}
    \stt{move(A, R, C, t) :- reachable(A, R, C, t).}
\end{equation*}
\stt{A, R, C, t} variables are leaf nodes for the corresponding predicate atoms \stt{move, reach\-able}. Moreover, \stt{reachable} is a leaf node for \stt{move}, after the implication operator.
Given such a tree structure, a solver for Prolog is able to respond to queries from the user implementing a backtracking algorithm. 

Starting from a user's query and available initial information (e.g., from sensors), Prolog solvers implement backtracking search along the tree branches, in order to verify whether atoms in the query can be grounded.
Moreover, last explored branches in the tree are saved in a stack, so as to retrieve the root when a branch returns no solution and to restart the search process. More details on the algorithm can be found in \cite{NarbG99} and related references. 
Researchers have tried to improve the efficiency of this solving procedure, e.g., with intelligent backtracking based on the generator/consumer paradigm \citep{KL88} or with parallel AND-tree verification in \cite{PMCA86}. More recently, \cite{ZH19} presented an approach based on linear model reduction.

Backtracking requires the prior stratification of the program in order to guarantee that solving terminates. Stratification guarantees that a (propositional) logic program can be partioned in incremental subsets of clauses $P_1 \cup ... \cup P_n$, such that:
\begin{itemize}
    \item if a predicate $p$ appears as positive in $P_i$, then it is defined in $\bigcup_{j\leq i} P_j$;
    \item if a predicate $p$ appears as negative in $P_i$, then it is defined in $\bigcup_{j< i} P_j$.
\end{itemize}
Since Prolog-based programs represent knowledge in a hierarchical tree structure, this subset composition guarantees that the tree search never gets stuck in a loop between predicates defined at different levels of the hierarchy. 
However, not all logic programs can be stratified, as evidenced by \cite{DrabW96}. 

\subsection{Planners based on the answer set semantics}\label{subsec:asp_review}
A non-monotonic logic programming language with similar syntax to Prolog is \emph{Answer Set Programming (ASP)} \citep{calimeri2020asp}. Introduced by \cite{MT99}, ASP was first proposed in \cite{LifsV99} to solve the planning problem for autonomous agents, as applied, e.g., by \cite{dimopoulos2019plasp}. 
ASP is based on the \emph{stable model semantics} \citep{GL88}. In order to understand what a stable model, or answer set \citep{LifsV99}, for a logic program is, consider a classical (i.e., NAF-free) propositional logic formula $F$ depending on a set $X$ of Boolean variables. The \emph{reduct} of $F$ to $X$, represented as $F^X$, is the formula resulting from the substitution of all variables appearing with negation with the false $\bot$. We then define a \emph{stable model} (or \emph{answer set}) of $F$ as a \emph{minimal model satisfying $F^X$}, namely a set $Y \subseteq X$ with no proper subsets satisfying $F^X$. For instance, let $F$ be:
\begin{equation*}
    (q \rightarrow p) \land (\lnot r \rightarrow q) \land (p \rightarrow s)
\end{equation*}
$Y = \{s, p, q\} \subset X = \{s, p, q, r\}$ is an answer set for $F$, since it satisfies $F^X : \ (q \rightarrow p) \land (\lnot \bot \rightarrow q) \land (p \rightarrow s)$ and no proper subset of $Y$ does. If $F$ includes NAFs, they are converted to classical negation (closed world assumption) and then the stable model is computed.

With reference to the peg transfer domain, consider the precondition for transfer action:
\begin{equation*}
    \stt{transfer(A, R, C1, t) :- holding(A, R, C1, t), not on(R, C1, P, C2, t).}
\end{equation*}
where \stt{not} is ASP NAF syntax, similar to Prolog.
In case sensors provide the initial grounding \{\stt{t = 1, A = psm1, C1 = red, holding(psm1, ring, red, 1)}\}, \stt{not on(R, C1, P, C2, t)} is converted to $\lnot$\stt{on(ring, red, peg, \_, 1)}, where \stt{\_} represents any possible assignment for \stt{C2}. Hence, the following propositional logic formula is generated:
\begin{equation*}
    \stt{transfer(psm1, ring, red, 1)} \leftarrow \stt{holding(psm1, ring, red, 1)} \land \lnot \bot
\end{equation*}
whose stable model is \{\stt{transfer(psm1, ring, red, 1), holding(psm1, ring, red, 1)}\}. In other words, the transfer action can be executed.
From this example, it is evident that an answer set includes \emph{all ground atoms}, i.e., not only actions, but also environmental fluents. This represents one main difference with respect to Prolog-based systems, which instead answer to user's specific queries. Computation of answer sets enhances the interpretability of the task planner, returning both the \emph{input} (environmental context) and \emph{output} (sequence of actions) to the task planning problem. 

In general, stable models are not unique. 
This is evident, for instance, considering the ASP program describing the peg transfer task. For instance, when multiple rings are present in the peg transfer domain, their order is arbitrary, so multiple plans can be generated, each having a different ordering of rings. In this case, each plan is part of a separate answer set for the logic program. 

ASP introduces additional constructs with respect to Prolog. 
Starting from the first solvers Smodels \citep{syrjanen2001smodels} and DLV \citep{eiter2000declarative}, up to the latest extensions as DLV2 \citep{adrian2018asp} and Clingo \citep{gebser2019multi}, it is possible to define \emph{aggregates} in the form \stt{l\{a : b\}u}.
An aggregate defines a set of atoms \stt{a}, subject to preconditions \stt{b}, such that the cardinality of the set (i.e., the number of grounded atoms \stt{a} in the answer set) is bounded within \stt{l} and \stt{u}). This construct translates to a \emph{choice rule}, which is useful to generate alternatives between sequences of actions in the task planning domain.
For instance, multiple rings may be present in the peg transfer scenario, and either starting from any of them is possible to complete the task.
However, it is possible to include preconditions to actions in an aggregate, e.g., for \stt{move(A, R, C, t)}:
\begin{equation*}
    \stt{0\{move(A, R, C, t) : reachable(A, R, C, t)\}1.}
\end{equation*}
In this way, up to only one action will be grounded in the answer set per time step, hence only one (random) plan will be finally generated. 
Similarly, it is possible to define an aggregate rule with an external condition:
\begin{equation*}
    \stt{0\{move(A, R, C, t) : reachable(A, R, C, t)\}1 :- arm(A).}
\end{equation*}
In this case, the solver will ground up to one action \emph{for each PSM}, so that the two robotic arms can operate at the same time on different rings, reducing the time for completing the task.

Choice rules may be supported by \emph{preference reasoning} in ASP \citep{BNT08} with \emph{weak constraints}, or corresponding optimization statements introduced by the state-of-the-art Clingo system \citep{gebser2019multi}.
A weak constraint is expressed with the following syntax:
\begin{equation*}
    \stt{:\texttildelow \ b}_{1..n}\stt{(V}_{1..m}\stt{).[w@p, V}_{1..m}\stt{]}
\end{equation*}
where \stt{b}$_{1..n}$ are body atoms depending on \stt{V}$_{1..m}$ variables, \stt{w} is the \emph{weight} and \stt{p} is the \emph{priority level}.
The meaning of above syntax is that simultaneous grounding of body atoms has a cost equal to the weight.
Answer sets \emph{must} then guarantee the satisfaction of hard constraints defined with the usual syntax \stt{:- atom}, but additionally shall \emph{preferrably} minimize the weight of the weak constraint. If multiple weak constraints are grounded, then the cumulative weight has to be maximized, but priority is given to constraints with lower \stt{p}.  
With reference to the example peg transfer task, assume to maximize economy of motion, e.g., moving first to the closest ring (to any PSM). This can be easily expressed introducing a new fluent in the domain description, \stt{distance(A, R, C, X, t)}, defining the distance between an arm and a ring as \stt{X}\footnote{\stt{X} variable can only have integer value in ASP. It is then important to convert actual distances between objects of the domain to integers, e.g., with ranking over all distances as shown by \cite{IROS2020}}. Then, it is possible to add the weak constraint:
\begin{equation*}
    \stt{:\texttildelow \ move(A, R, C, t), distance(A, R, C, X, t).[X@1, A, R, C, X, t]}
\end{equation*} 
in order to minimize the distance of target ring.
Equivalent Clingo syntax for the weak constraint notation is the optimization statement:
\begin{equation*}
    \stt{\#minimize\{X: move(A, R, C, t) : distance(A, R, C, X, t)\}}. 
\end{equation*}

ASP solvers compute answer sets with methods inherited from the area of satisfiability (SAT) checking \citep{biere2009handbook}. The basic approach consists in bottom-up grounding of initially known atoms (e.g., environmental information from sensors), and incrementally verifying logic implications.
It is important here to mention Clingo, which allows to structure the ASP program in parts or sub-programs, specifically \emph{base, step} and \emph{check}. In a task planning domain, the base part typically contains definitions of resources, while specifications are in the step part and the goal is in the check sub-program. The solver verifies at each time step the check part, and increments the temporal variable to ground specifications only if the goal is not satisfied. In this way, the computation of the plan continues \emph{until the goal is not satisfied}, thus implementing some capabilities of a temporal logic formalism.

Differently from Prolog tree search based mainly on Selective Linear Definite with Negation as Failure (SLDNF) \citep{AD94}, ASP solvers guarantee termination \citep{LifsV08}. Moreover, SAT solving has applications in a variety of fields of computer science. For this reason, starting from the popular MiniSAT \citep{sorensson2005minisat}, a lot of research has been devoted to improve the efficiency of existing SAT solvers (inherently NP complete) even in complex domains (see \cite{manthey2016towards} for recent advances), e.g., with parallel evaluation of specifications as in Clingo \citep{kaufmann2016grounding}. Furthermore, ASP syntax forces the user to define \emph{safe statements}, i.e., all variables in the head must appear also in the body without NAF. In this way, propagation of grounded atoms from bodies to heads is more efficient, since variables in the head inherit a specific assignment from the body.

Though a multitude of applications of ASP can be found in AI planning (see \cite{erdem2016applications} for a recent review), the applications in robotics are still fewer than Prolog-based systems, which have been well established for a significantly longer time. This is also due to the development of research for efficient SAT solving algorithms, which is significantly advancing only in recent years. 
Currently, the most popular implementation for ASP, with applications in the robotic scenario, is Clingo, which from the latest release (Clingo 5) also implements constraints on reals. 

Relevant applications of Clingo have been proposed by \cite{PSTY19}, where its variant Asprilo \citep{gebser2018experimenting} for multi-robot logistics and warehouse automation, is exploited; \cite{OJG18}, where Clingo system is integrated with the ALICA language for teamworking in domestic and general-purpose environment \citep{OJJG19}; \cite{cui2021} for general-purpose service robotics; \cite{lu2017integrating}, proposing a task planning framework for domestic service robots receiving information and goal description in natural language from humans, at the RoboCup@Home challenge; \cite{IROS2020, meli2021autonomous, tagliabue2022deliberation}, where an integrated framework for interpretable surgical task-motion planning is proposed; \cite{andres2015integrating}, which shows the ROSoClingo package integrating Clingo in ROS. A recent application based on the DLV system can be found in \cite{WZZQG09}, addressing the problem of mechanical assembly sequence in industry. 

Several other implementations of ASP have been proposed by \cite{tu2007reasoning}, where an Answer Set-based Conditional Planner (ASCP) is presented with applications also to robotic examples; \cite{chen2013toward}, where ASP is implemented on the OK-KeJia service robot prototype for planning and decision-making based on multiple sources of information in the context of human-robot interaction; \cite{baral2015add}, proving the advantages of using ASP for task planning and failure analysis and explanation in human-robot interaction; \cite{ji2009cognitive} for the domestic service scenario; \cite{SGZW19}, combining ASP with Markov decision processes in a domestic scenario; \cite{bertolucci2021manipulation} for dual arm manipulation; \cite{FFSTT18} propose a review of recent applications of ASP to industry, while \cite{erdem2018applications} survey recent applications in various robotic fields.
Finally, we mention recently growing interest in ASP applications to the scheduling problem, either in transportation / routing \citep{gebser2018routing, abels2019train} or space assignment \citep{dodaro2021asp, dodaro2022operating} scenarios. Though robotic implementations are still missing, we believe this is an interesting application of ASP to trending robotic problems, e.g., autonomous vehicle and industrial / service automation design.

\subsection{Other planners}
\label{subsec:others}
While Prolog- and ASP-based planners are most popular and can be applied to a wide range of domains, several other implementations of planners based on standard logic programming exist.
An example are constraint-handling rules \citep{fruhwirth1998theory}, which improve performance of the solving process simplifying task-specific constraints online. An example of this approach is implemented in fluent executor \citep{Thie05}, which has been used, e.g., for the automation of an office robot, hence in an application involving human-robot interaction \citep{wu2008office}. 

Analogous application-specific approaches rely directly on a generic description of the task in an action language (e.g., PDDL), and implement a custom solution approach for the specific use case.
An important example of this class of planners is ROSPlan \citep{cashmore2015rosplan}, a framework combining PDDL-based task planning, motion planning, sensing and open-world planning. 
Basing on ROS infrastructure offers a considerable implementative advantage for application with most robotic systems.
ROSPlan has been applied in several scenarios, e.g., by \cite{miranda2018rosplan} for multi-robot navigation; \cite{sanelli2017short}, integrated with Petri nets for human-robot interaction; \cite{EA16}, tackling the anchoring problem, that is matching abstract symbols from high-level knowledge to perceptions and executed actions, in order to automatically generate code for ROS nodes.

Other examples of custom implementations derived from PDDL are based on action graphs \citep{HS20}. \cite{gragera2019modelling} proposed a PDDL-based planner for social robotics; \cite{ML19} investigated coordination of unnamed ground and aerial vehicles; \cite{MRBR19} implemented an integrated framework for robotic manipulator, surveillance and rover exploration; \cite{MFM16} focused on mobile robots in the ROS framework; SHOP \cite{nau1999shop} was proposed for a state-of-the-art implementation of Hierarchical Task Network (HTN) \citep{erol1996hierarchical}, a popular and efficient formalism for task planning with sub-goals. 

It is important here to highlight that task-specific approaches derived from generic PDDL descriptions of the task are not the only choice, and other custom implementations based, e.g., on ASP- and Prolog-like syntax are possible and can be adapted to the specific needs of the use case.
For instance, in \cite{dix2003planning} AnsProlog, the language for answer set semantics, was used to implement HTN, proving comparable efficiency with respect to SHOP (a more extended comparison is available in \cite{son2006domain}). Moreover, recently PDDL- and ASP-based planners have been compared by \cite{JZKS19}. The authors compare Clingo as the state-of-the-art implementation of ASP against the two most prominent FastDownward PDDL-based planners FDSS-1 \citep{helmert2011fast} and LAMA-2011 \citep{richter2011lama}, in different experimental scenarios including robot navigation. The comparison shows that PDDL-based systems are generally more efficient when the required plan for the task is composed of a long sequence of actions. However, the computational performance of ASP-based systems are better for shorter plans, and they scale significantly better with the size of the domain description (i.e., the number of resources and specifications).

\subsection{Discussion}
\label{subsec:std_disc}
Prolog- and ASP-based systems are the most prominent solutions to the task planning problem in robotics. Their main advantage lies in the non-monotonic representation of task knowledge, which is essential for robotic applications since it allows plan revision when environmental conditions change. Prolog-based planners as SWI-Prolog \citep{WSTL12} are query-based planners which also support advanced reasoning on continuous variables, bridging the gap between the discrete logic representation of the robotic scenario and the continuous perception of, and interaction with, the environment. However, this may reduce the computational efficiency of solving the planning problem online \citep{helmert2002decidability}. Instead, ASP-based frameworks as Clingo by \cite{gebser2019multi} directly compute more interpretable and informative \emph{answer sets}, i.e., sets of ground atoms representing the current environmental conditions and the possible plans to accomplish a given task. Moreover, they allow multiple plan generation and preference reasoning for optimal planning. 

All standard logic programming frameworks allow to represent temporal concepts (e.g., temporal sequence, or event-driven properties) by introducing an additional variable for temporal indexing, as shown in the beginning of this section and by \cite{son2006domain, meli2021inductive}.
In particular, examples of temporal Prolog extensions can be found in \cite{abadi1989temporal}, while recently temporal extensions to the ASP syntax \emph{telingo} \citep{cabalar2019telingo} and \emph{tasplan} \citep{cabalar2019complete} have been proposed, based on Clingo solver and temporal equilibrium logic \citep{aguado2013temporal}. Although no relevant robotic applications of these paradigms are known to the authors, temporal extensions of ASP and Prolog may well find a wide range of applications in the future, since they offer invaluable advantages, e.g., non-monotonicity in the task knowledge representation.

For these reasons, the research community has put considerable effort in implementing time- and memory-efficient solvers, especially in the field of SAT solving for ASP-based planners. This balances the increase in the size of represented knowledge and computational cost, e.g., when additional variables and specifications for temporal reasoning are included. Also, state-of-the-art frameworks as Clingo and SWI-Prolog offer C / C++, Java and Python APIs, as well as easy integration with ROS.

In addition, some standard logic programming languages can support the open-world assumption, integrating external knowledge bases to be queried when needed during the plan execution, in order to implement modular and more efficient knowledge representation in very complex domains. 
In fact, the knowledge base encodes a detailed description of the scenario, while the logic program only encodes task specifications and queries the knowledge base for selective grounding.
The most popular example of this approach is the CRAM framework by \cite{beetz2010cram} for advanced robotic manipulation, based on Knowrob \citep{TB17} ontology for general service robotic tasks. Knowledge is queried through SWI-Prolog, as validated, e.g., by \cite{YBHBBB19}, where an outdoor search-and-rescue robotic task is fulfilled, performing geometric reasoning and semantic modeling of the environment; \cite{CBMBB18} for multi-objective navigation tasks; \cite{BPB18} for incomplete assembly task in industry; \cite{GT15} for task planning in orthopaedic surgery, integrating an orthopaedic-specific knowledge base; \cite{PSL15} for the safety-critical scenario of mine-countermeasures missions for autonomous underwater vehicles, addressing the challenge of recovery from hardware failure and changes in priority levels. 
Another implementation, based on the answer set semantics, is a recent extension of DLV, DLVHEX \citep{eiter2018dlvhex}, which provides specific constructs to query external knowledge bases. 
While not providing the same expressive power as description logics, the query-based approach integrating logic programming and ontologies is usually more efficient and preserves the non-monotonic assumption, which is essential in most applications.

\section{Temporal logic programming}\label{sec:temporal}
Standard logic programming offers the expressivity to describe many task planning problems in robotics, with operators representing causal relations between atoms. However, in several scenarios additional expressivity is required to describe complex temporal relations between resources. 
The scope of this section is to present logic programming solutions based on temporal logic formalism. Main temporal operators for robotic applications are: 
\begin{itemize}
    \item \emph{eventually} $\lozenge$\stt{a}, prescribing that a statement or atom \stt{a} shall hold at some point in the future;
    \item \emph{next} $\ocircle$\stt{a}, specifying that \stt{a} holds at the subsequent time step;
    \item \emph{always} $\square$\stt{a}, prescribing that \stt{a} can never be false;
    \item \emph{until} \stt{a}$\pazocal{U}$\stt{b}, meaning that \stt{a} must hold at least until \stt{b} becomes true;
    \item \emph{release} \stt{a}$\pazocal{R}$\stt{b}, specifying that \stt{b} holds until \stt{a} becomes true (in this sense, \stt{a} \emph{releases} \stt{b}).
\end{itemize}
As mentioned in the previous section, some of these temporal relations (e.g., effects of actions with temporal delay) can be represented even with standard logic formalism. However, this requires the definition of an explicit temporal variable \stt{t}. Moreover, it is not straightforward to express complex relations without specific temporal operators. For instance, consider the peg transfer task, and assume that rings have a significant mass, so that they exert high force on the grasping tool. In this case, a reasonable approach is to slowly lift a ring, e.g., to extract it from a peg, while continuously monitoring the force sensed at the end effector. If some safety threshold is overcome, then extraction must be interrupted and a recovery policy is needed.   
Using standard Boolean logic, e.g., with ASP, this could be expressed with the following set of statements:
\begin{align*}
    \stt{extract(A, R, C1, t) } &\stt{:- holding(A, R, C1, t), on(R, C1, P, C2, t)}\\
    \stt{extract(A, R, C1, t) } &\stt{:- extract(A, R, C1, t-1), not max\_force(A, t),}\\
    &\ \ \ \ \ \ \stt{on(R, C1, P, C2, t)}
\end{align*}
where \stt{max\textunderscore force(A, t)} is a fluent grounded when force sensors detect high force at the tool. The two above statements can be replaced with a single statement in temporal logic, with the $\pazocal{R}$ operator:
\begin{align*}
    \stt{not on(R, C1, P, C2) } \pazocal{R} \ (\stt{extract(A, R, C) }&\leftarrow \ \stt{holding(A, R, C),}\\
    &\ \ \ \ \ \ \ \stt{not max\_force(A)})
\end{align*}

Logic programming framework in this section mainly rely on linear temporal logic (LTL) by \cite{PnueA77}, assuming a linear sequence of events, rather than multiple realizations as in more complex (and computationally inconvenient) non-linear temporal logic (e.g., CTL by \cite{emerson1982using}). Specifically, an interpretation of LTL is considered over \emph{finite traces} (LTL$_f$), i.e., over a finite time horizon, as proposed, e.g., by \cite{bienvenu2006planning,gabaldon2004precondition,wilke1999classifying}. 
This is a reasonable assumption for robotic tasks, which typically involve a limited sequence of actions to be performed. The choice to focus on LTL is justified by two reasons. First, LTL$_f$ has been proved to be as expressive as first order logic in \cite{diekert2008first}. LTL's extension LDL$_f$ \citep{de2013linear}, including regular expressions from propositional dynamic logic \citep{fischer1979propositional}, has been proved to be as expressive as monadic second order logic for finite automata by \cite{trakhtenbrot1962finite}. The problem of synthesis for LDL$_f$ formulas (i.e., the assignment of truth values to variables so that the given formulas hold) has been proved to be equivalent to the problem of conditional planning under full observability, and 2EXPTIME-complete \citep{de2015synthesis}. Hence, it is possible to encode a generic task planning problem from high-level action languages (e.g., PDDL) into LTL$_f$/LDL$_f$ formulas, which are then solved through methods based on SAT checking. 
For simplicity, the subscript $f$ will be omitted through the rest of this section, hence LTL will stand for LTL$_f$. 

A relevant tool for LTL in the robotic community is LTLMoP by \cite{finucane2010ltlmop}, to implement temporal task specifications in the ROS framework. LTLMoP has been used for multi-target navigation \citep{kumar2016linear}; task planning in a grocery store with an Aldebaran Nao robot \citep{jing2012correct}; multi-robot cooperative task planning \citep{kim2012simulation,KK14}; search-and-rescue mission in partially known environment with a Pioneer 3-DX robot \citep{sarid2013guaranteeing}; task planning for modular robots \citep{jing2016end}; multi-task planning for a patrol robot \citep{LPZCYZ17}; reactive task planning based on game theory \citep{SML17}. 
Also relevant is the application of LTLMoP to the problem of explanation of unsynthetizable plans for robotics \citep{raman2012explaining}, which is useful to enhance the interpretability of the robotic system. 
LTLMoP implements SAT checking to solve the task planning problem from temporal specifications.
As evidenced by \cite{kress2011correct}, one limitation of this approach is the \emph{state explosion} problem, which is caused by the introduction of the temporal dimension in the planning problem and hence the different time horizons to be evaluated at the solving stage. This becomes even more crucial when implementing LTL for on real robots, typically with hybrid control (connecting discrete-time and continuous-time control) so that the lower-level motion behavior matches the task-level specifications. In this scenario, LTL-based plan synthesis becomes often intractable, due to infeasibility of analyzing all future scenarios. In order to cope with the computational issues, several solutions have been proposed. 

For instance, in \cite{LAFKV15} the concept of \emph{quantification of satisfiability} has been introduced, in order to guarantee the satisfaction of main LTL constraints while neglecting less important ones. This solution is applied, e.g., in \cite{TMDK14}, where maximal satisfaction of LTL task specifications is guaranted for a Nao robot able to grasp, drop a ball and walk managing failure conditions. 

Another widely used approach is the \emph{receding time horizon} \citep{wongpiromsarn2010receding}, which relies on the assumption that the dynamic nature of the environment does not allow to devise a plan which extends long ahead in the future. Hence, a good design choice is to verify LTL formulas with a short time horizon, and to refine the resulting plan on the fly during the execution, according to the current state. In this way, only most probable future traces are analyzed, and the state explosion is significantly mitigated.
A tool which implements the receding time horizon approach, differently from LTLMoP, is TuLiP \citep{WTOXM11}. Though it has been mostly applied to the provably-correct construction of hybrid controllers for integrated task-level (discrete) and motion planning (continuous) problems, worth mentioning are the applications in \cite{lin2014mission}, where a robot surveillance task is planned through TuLiP, while local reactivity as in \cite{bloem2012synthesis} and refinement is guaranteed through $\mu$-calculus, a modal logic which can be used to express temporal formulas \citep{bradfield2007modal}; and \cite{zhao2016high} for whole-body biped locomotion in unstructured environment.

The two above solutions are unfortunately not always applicable in robotics. In fact, quantifying the relevance of constraints to be considered in the solving process is dangerous in safety-critic scenarios, since it could lead to undesired behavior. As for limiting the maximum time horizon, it is possible for simple tasks as the peg transfer, but it does not scale properly to complex tasks with long-term goals, e.g., exploration of wide and possibly unknown environments.

Other LTL-based paradigms for robotic task planning exist, which have the same issues as the above mentioned ones, though. They find mostly application in multi-robot systems coordination \citep{KB06,WK16,LL12}, with Petri-net task representation; \cite{GTD14}, with dynamic leader selection under communication constraints; \cite{GTD16} with mutual distance constraints; \cite{KFP09} for search-and-rescue and coverage scenarios; \cite{HLKV15} for manipulation tasks; \cite{SBD17} for time optimality under the ROS framework; \cite{9372768}, where signal temporal logic \citep{donze2010robust}, a temporal logic devised for motion planning, is combined with standard LTL to deal with event-based task planning in a multi-robot domain. It is also important to mention other applications, including \cite{XLNS16} where dynamic planning in a robotic fire-fighting scenario is investigated; \cite{CMO17} where the construction of behavioral trees from LTL specifications is used for simulated task planning of a mobile robot in adversarial environment; \cite{BDR18} for autonomous flexible manufacturing with experimental evaluation on a Gantry robot in Siemens NX Mechatronics Concept Designer simulator; \cite{HHWZZPY17} for warehouse robotics; \cite{GJD13} for robot navigation, integrated with lower-level model-ckecking-based motion planning through hybrid control;  \cite{CTMBM17}, where LTL over finite and infinite traces specifications derived from PDDL encoding are solved through translation to non-deterministic automata; \cite{choi2021}, where LTL is used to design a medical robot tasked with roles of patient reception and triage. 

\subsection{Discussion}
Temporal logic programs can represent complex temporal task specifications in an elegant formalism, without the need to use an ad-hoc variable for temporal indexing. This is particularly useful in applications concerning involving long planning horizon, or when formal properties of the robotic system and the environment must be guaranteed (e.g., stability or reachability of safe states in integrated task-motion planning). 
The need for considerable temporal expressivity is evidenced by the many applications of temporal extensions of PDDL implemented in LTL syntax, e.g., in industrial automation \citep{kootbally2016industrial} or for multi-robot mission planning \citep{wurm2010coordinated, wurm2013coordinating, coles2019board, carreno2020decentralised}.

One main limitation of temporal logic programming is the non-monotonicity of most formalisms (apart from few exceptions with no application to robotics, e.g., by \cite{baral2007non}), which hinders online reconfigurability and plan revision in under dynamic environmental conditions.
Moreover, the increased expressive power brings an added computational cost, induced by the inherent complexity of the decidability problem for temporal planning \citep{rintanen2007complexity}. For this reason, several temporal extensions of standard logic programming paradigms exist, e.g., Prolog-like \citep{abadi1989temporal} and more recent ASP-based \emph{telingo} by \cite{cabalar2019telingo} mentioned in Section \ref{subsec:std_disc}.

Nevertheless, recently a significant effort in implementing efficient LTL SAT checking is being devoted, e.g., by \cite{dureja2018more, geatti2021black}, in order to bridge the gap between temporal logic programming and practical robotic use cases. 
Main paradigms presented in this section, i.e., TuLiP \citep{WTOX11} and LTLMoP \citep{finucane2010ltlmop}, also offer implementation support via, e.g., Python APIs, and the latter has been already integrated in the ROS framework.

\section{Probabilistic logic programming}\label{sec:probabilistic}

The logic programming solutions presented in the previous sections allow to describe task planning problems with complex constraints, involving temporal and causal relations. However, an important challenge for deliberative robots interacting with unknown environment is the quality of the perception. This requires a real-time planning system which not only accounts for the real-time information from sensors to update the environmental model, but also considers uncertainty of the sensed model and generates a sequence of \emph{most probable} actions, to be refined as new evidence is acquired. In this section, logic programming languages are analyzed which enrich standard and temporal logic semantics with concepts from probability theory, in order to express specifications with some degree of uncertainty. Probabilistic logic definitions can be found, e.g., in \cite{NS92}, Distribution Semantics (DS) in \cite{SatoT95} and relative extensions \citep{LukaT98, LukaT01}. 

The syntax of probabilistic logic programs includes p-annotated atoms and rules. Given $\mu \subseteq [0,1]$, a \emph{p-annotated Boolean} $a:\mu$ is a Boolean variable which is true with probability in $\mu$. Similarly, in the formalism of logic programming it is possible to define p-atoms which can be grounded with bounded probability. A \emph{p-statement} is then a statement either containing p-annotated atoms, or with its own annotation (i.e., $(a \leftarrow b):\mu$).
For instance, in the peg transfer domain the location of a ring may be known with accuracy between 50\% and 80\%, depending on noisy point cloud from the RGBD camera. This affects the probability of all atoms related to ring's position, e.g., \stt{reachable(A, R, C, t)} becomes a p-atom \stt{reachable(A, R, C, t):}$[0.5, 0.8]$. 
As a consequence, all statements including the reachability atoms become p-statements. For example, the precondition for moving to a ring becomes:
\begin{equation*}
    \stt{move(A, R, C, t)} \leftarrow \stt{reachable(A, R, C, t)}:[0.5, 0.8]
\end{equation*}
Also \stt{move(A, R, t)} then becomes a p-annotated atom, which inherits the same probability range as the p-atom precondition. If multiple preconditions hold for an atom, the annotations of all of them concur at defining its probability range. 

As a consequence, the solving process for probabilistic logic programs suffers from a state explosion problem similar to temporal logic. In fact, it is not possible to ground an atom as true or false (assuming closed-world assumption), unless the probability range reduces to $\{0\}$. Then, grounding atoms according to specifications involves combining different probability ranges for body atoms, and then propagating them to head atoms, evaluating all possible scenarios.

Several preliminary solutions have been proposed to make probabilistic logic solving more efficient. For instance, it is possible to restrict the solution of the logic program to the deduction of only tight logic consequences. Considering a set of probabilistic propositional logic clauses (resulting from grounding of p-statements) $\pazocal{P}$, a clause \stt{c:}$\mu_c$ is said to be a \emph{tight logic consequence} of $\pazocal{P}$ if $sup \mu_c = max \mu, inf \mu_c = min \mu$, where $\mu \subseteq [0,1]$ is the set of real numbers for which there always exists a realization for clauses in $\pazocal{P}$. In other words, a clause is a tight logic consequence of a set of clauses if it can be deduced from all their possible realizations. 
As an example, consider the peg transfer domain. From specifications, we know that it is possible to place a ring on a peg either by directly moving it to the peg, or by transferring it to the other arm first, depending on reachability conditions. Assuming that environmental fluents are p-atoms due to sensor uncertainty, we can encode this knowledge with the following probabilistic clauses:
\begin{align*}
    (\stt{on(ring, red, peg, red, 2)} &\leftarrow \stt{move(psm1, peg, red, 1)}):[0.7, 1]\\
    (\stt{on(ring, red, peg, red, 3)} &\leftarrow \stt{move(psm1, peg, red, 2),}\\
    &\ \ \ \ \ \ \stt{transfer(psm2, ring, red, 1)}):[0.8, 1]
\end{align*}
and \stt{on(ring, red, peg, red, \_)}$:[0.7, 1]$ is a tight logic consequence. 

\cite{LukaT98} showed that the deduction of tight logic consequences can be reduced to solving two linear programs, assuming no negations are involved. However, the problem is shown to be NP in general, depending on the number of possible realizations. Hence, the author also proposed an algorithm to further mitigate the complexity of the problem when probabilistic clauses are separable from deterministic ones. 

Another solution consists in using approximate sampling methods for stochastic systems, e.g., Monte Carlo. This is used, e.g., in Problog \citep{de2007problog}, a popular probabilistic extension to Prolog which allows to define a weight assigning discrete probabilities to atoms and specifications. 
In this way, discrete probabilities are assigned to p-atoms and p-statements, instead of real ranges, and propagation of grounded atoms is more efficient (recall that reasoning on continuous variables has been proved in general undecidable by \cite{helmert2002decidability}).
Problog and its variants have been applied to several robotic task planning scenarios, e.g., for opening doors \citep{MAH13} and object manipulation \citep{moldovan2012learning, tan2019s, antanas2019semantic}. An alternative approach to Problog is the framework of distributional clauses (DC) by \cite{gutmann2011magic}, which extends Problog allowing to define continuous probability distributions over rules and atoms. DC has been used for robotic manipulation and grasping in \cite{moldovan2014learning, nitti2015planning, moldovan2018relational}. 

The above mentioned approaches are still approximate or can be applied only to a restricted class of logic programs, which makes them unsuitable for most real-time robotic domains. Nevertheless, probabilistic logic programming has the potential to fill the gap between classical logic determinism and the intrinsic uncertainty deriving from the planning-acting-sensing loop of deliberative robots. 

Recent research is in fact focusing on integrating query-based planners with probabilistic knowledge in open-world domains, which can be updated as new evidence is acquired (similarly, e.g., to what is done with the CRAM framework in standard logic). 
A relevant example is proposed in \cite{jain2009equipping}, where the ProbCog system allows to query and reason on a Bayesian logic network (BLN) derived from the probabilistic logic of multi-entity Bayesian networks (MEBN) \citep{laskey2008mebn} in a domestic scenario with a B21 service robot. Basically, a MEBN is a knowledge representation formalism which connects fragments of Bayesian networks through statements in first-order logic. This framework has been used, e.g., in \cite{beetz2012cognition} for complex task decomposition and planning with a robot for home chore; \cite{beetz2010towards, nyga2012everything} for manipulation; \cite{maurelli2014cognitive} for cognitive underwater vehicles. 

A similar approach is presented in \cite{nyga2018grounding} for domestic tasks, using a Markov logic network (MLN) \citep{richardson2006markov} instead of a BLN. MLNs are based on probabilistic weighted logic statements. In \cite{al2016robot} task planning for manipulation and navigation scenarios is implemented querying a MLN with the STRIPS-based Metric-FF planner \citep{hoffmann2001ff}; in \cite{lisca2015towards} a robot for chemical experiment automation is proposed querying probabilistic action cores (PRACs) \citep{nyga2012everything} based on MLNs. A mention is also deserved by LP$^{MLN}$ \citep{lee2016weighted, lee2018probabilistic} to reason on MLNs using the answer set semantics under the action language $\pazocal{BC}+$ \citep{babb2015action} for transition systems, though the actual application in robotic scenarios is still part of ongoing research.

It is also important to highlight that research is pushing towards the extension of the answer set semantics with probability theory. The main example is P-log \citep{baral2004probabilistic} which is based on weighted specifications as Problog, used, e.g., in the LCORPP framework for sequential robot planning \citep{amiri2019learning}. In this way, in the next future it may be possible to exploit recent advances in SAT solving strategies in the field of probabilistic logic.

Among other implementations of probabilistic logic programming with application to robotic scenarios, we mention PREGO \citep{belle2014prego}, which is similar to DC; Probabilistic Computational Tree Logic (PCTL) by \cite{rutten2004mathematical}, a probabilistic non-linear temporal logic, to define task specification, e.g., in human-robot interaction for manufacturing \citep{ZZL16} and robotic swarm coordination \citep{BPBD12}; \cite{GCMS14}, where event calculus based on MLNs is applied to a search-and-rescue scenario.

\subsection{Discussion}
Planners based on probabilistic logic programming allow to elegantly model uncertainty of real robotic scenarios into task specifications, extending rules and atoms with discrete probability values as in Problog by \cite{de2007problog}, or even with continuous distributions as in DC \citep{gutmann2011magic}. This allows to enrich the expressivity of purely probabilistic Markov or Bayesian networks, enhancing interpretability. The main representative tool, Problog, is an extension of Prolog, so it offers most of its advantages, including Python / Java APIs, advanced constructs and possibly integration with external knowledge bases for open-world reasoning as in CRAM \citep{beetz2010cram}. Similarly, probabilistic extensions have been proposed for temporal logic programming, e.g., PCTL by \cite{rutten2004mathematical}, and the answer set semantics \citep{baral2004probabilistic}. 

While several robotic applications of the probabilistic logic programming paradigms can be found in recent literature, the main limitation of current representation and reasoning tools is the computational inefficiency rising from the complexity of the planning problem under uncertainty \citep{helmert2002decidability}, especially when continuous (more realistic) probability distributions are modeled. 
As a consequence, the most efficient solution for combining advantages of formal logics and probability theory currently consists of a hierarchical composition of standard logic (e.g., ASP by \cite{SGZW19, leonetti2016synthesis} for autonomous navigation and exploration) or temporal logic programming \citep{leonetti2012automatic} with Markov models, in order to constrain probabilistic reasoning and generate more efficient and interpretable robust plans.

\section{Conclusion}
\label{sec:conclusion}
Standard logic programming frameworks are currently the most prominent solutions (in the field of logic-based approaches) to the task planning problem for deliberative robots, since they offer sufficient expressivity for most real use cases, sufficient software support for integration in robotic applications, and implement efficient solving algorithms. Prolog-based frameworks have the longest history and popularity, but ASP-based frameworks are rapidly emerging as valid competitors. The state-of-the-art solution in the answer set semantics is Clingo, which supports optimal plan generation through preference reasoning. In general, Prolog-based tools are designed to answer external queries, so they are best suited when reasoning on large knowledge bases (e.g., in the CRAM project) or in tight human-robot interaction. On the contrary, ASP-based planners should be preferred for advanced policy search in complex scenarios. 

Temporal logic provides a useful formalism for integrated task and motion planning (e.g., with TuLiP), hence it could prompt the development of next-generation deliberative robots.
Similarly, probabilistic logic promises to bridge the gap between planning and sensing.
Unfortunately, both formalisms suffer from computational inefficiency in realistic scenarios, due to state explosion problems. As a consequence, a generic framework has not emerged yet in the robotic community, and custom application-specific implementations, or alternative frameworks (e.g., Bayesian) are still the state of the art.

We believe that future research should push towards the integration of different frameworks, to find the best tradeoff between efficiency and expressivity.
For instance, a recent framework which proposes temporal ASP is \emph{telingo} by \cite{cabalar2019telingo}, while the hierarchical combination of logic planning and Markov decision processes has been proposed by \cite{SGZW19}.

\section{Declaratons}
\subsection{Funding}
This work was supported in part by the European Union Grant ERC-ADG N. 742671, ARS (Autonomous Robotic Surgery) project. URL: \url{https://www.ars-project.eu/?lang=it}

\subsection{Conflicts of interest/Competing interests}
The authors certify that they have no affiliations with or involvement in any
organization or entity with any financial or non-financial interest in
the subject matter or materials discussed in this manuscript.

\subsection{Availability of data and material}
Not applicable.

\subsection{Code availability}
Not applicable.

\bibliographystyle{spbasic}      
\bibliography{biblio}   
\end{document}